\documentclass[11pt,a4paper]{article}
\usepackage{times,latexsym}
\usepackage{url}
\usepackage[T1]{fontenc}

\usepackage[acceptedWithA]{tacl2018v2}

\usepackage{xspace,mfirstuc,tabulary}

\newif\iftaclinstructions
\taclinstructionsfalse 
\iftaclinstructions
\renewcommand{\confidential}{}
\renewcommand{\anonsubtext}{(No author info supplied here, for consistency with
TACL-submission anonymization requirements)}
\newcommand{\instr}
\fi

\iftaclpubformat 

\else

\fi

\title{ \vspace*{-0.5in}
{{\small \hfill TACL'21}\\
\vspace*{.25in}} Towards Question-Answering as an Automatic Metric for Evaluating the Content Quality of a Summary}

\author{
 Daniel Deutsch,$^{\dagger}$
 Tania Bedrax-Weiss,$^{\textrm{\textdaggerdbl}}$
 and
 Dan Roth$^{\dagger}$\\
 $^\dagger$Department of Computer and Information Science, University of Pennsylvania \\
 $^{\textrm{\textdaggerdbl}}$Google Research \\
 \texttt{\{ddeutsch,danroth\}@seas.upenn.edu} \\
 \texttt{tbedrax@google.com}
}

\usepackage{microtype}
\usepackage{booktabs}
\usepackage{graphicx}
\usepackage{adjustbox}
\usepackage{multirow}
\usepackage{subcaption}  
\usepackage[font=small]{caption}  
\usepackage{amsmath}
\usepackage{makecell}  
\usepackage{resizegather} 

\date{}

\begin{document}
\maketitle

\begin{abstract}
A desirable property of a reference-based evaluation metric that measures the content quality of a summary is that it should estimate how much information that summary has in common with a reference.
Traditional text overlap based metrics such as ROUGE fail to achieve this because they are limited to matching tokens, either lexically or via embeddings.
In this work, we propose a metric to evaluate the content quality of a summary using question-answering (QA).
QA-based methods directly measure a summary's information overlap with a reference, making them fundamentally different than text overlap metrics.
We demonstrate the experimental benefits of QA-based metrics through an analysis of our proposed metric, QA\-Eval.
QA\-Eval out-performs current state-of-the-art metrics on most evaluations using benchmark datasets, while being competitive on others due to limitations of state-of-the-art models.
Through a careful analysis of each component of QA\-Eval, we identify its performance bottlenecks and estimate that its potential upper-bound performance surpasses all other automatic metrics, approaching that of the gold-standard Pyramid Method.\footnote{
Code is available at \url{https://github.com/CogComp/qaeval-experiments}.
}

\end{abstract}
\section{Introduction}
Evaluating the content quality of a summary is a fundamental task of text summarization.
As such, it has received the attention of researchers for the past two decades \citep[][\emph{i.a.}]{Lin2004,Nenkova2004,Hovy2006,Louis2013,Zhao2019}.
The most popular approaches are reference-based metrics, which treat a human-written reference summary as the gold standard and score a candidate summary based on how similar its content is to the reference.

It is desirable to have reference-based evaluation metrics that calculate this similarity score based on how much information the two summaries have in common.
The vast majority of previous automatic evaluation metrics compare two summaries based on matching their tokens, either through some lexical \citep{Lin2004,Hovy2006,Tratz2008} or embedding-based similarity \citep{Zhang2019,Zhao2019}.
Although they capture a valuable quality signal, these methods match tokens which do not express the same information and instead end up comparing the similarity of two summaries based on the topics they discuss \citep{Deutsch2020c}.

In this work, we propose a metric to evaluate the content quality of a summary using question-answering (QA).
Metrics within a QA evaluation framework represent the information of a reference summary using QA pairs, then estimate how much of this information is contained in a candidate summary by calculating the proportion of questions it can answer.
Because the questions can only be answered if the candidate summary contains the corresponding information, QA-based metrics directly measure the information overlap, providing a summary quality signal that is not effectively captured by text overlap based metrics.

We build upon previous work in this direction \citep{Eyal2019} and propose and analyze a more general QA-based metric, which we call QA\-Eval (\S\ref{sec:pipeline}).
We experimentally show the benefit of QA\-Eval, both with current state-of-the-art methods and by estimating its potential upper-bound performance.

We show that with current question-generation and question-answering models, QA\-Eval achieves state-of-the-art correlations to human judgments on benchmark datasets when used to evaluate summarization systems (by averaging scores over dozens of summaries), outperforming all other automatic metrics and equalling the gold-standard Pyramid Method \citep[][\S\ref{sec:overall}]{Nenkova2004}.
When used to rank individual summaries, the metric is equal or better to other metrics on summaries that are very similar to the ground-truth and is competitive on others due to shortcomings of current state-of-the-art models (\S\ref{sec:question_answering}).

Through a careful analysis of each component of QAEval (\S\ref{sec:answer_selection}-\S\ref{sec:question_answering}), we identify both the QA model and verifying if the predicted answer is correct as the performance bottlenecks (\S\ref{sec:question_answering}), whose noise likely explains the lower summary-level performance in some scenarios.
Based on a manually annotated set of 2.9k QA pairs, we show that with human-level QA and answer verification performance, the summary-level upper-bound correlations of QA\-Eval are better than all other automatic metrics and approach the gold-standard Pyramid Method.
In combination with state-of-the-art correlation results, this strongly indicates that QA-based evaluation metrics are a promising direction for future research.

The contributions of this work include
(1) a proposal of QAEval, a more general QA-based metric for evaluating the content of summaries,
(2) experimental evidence that demonstrates QAEval's state-of-the-art performance on benchmark datasets, 
(3) an analysis that identifies the QA model and answer verification as the performance bottlenecks,
and (4) an estimate that QAEval's upper-bound summary-level performance in scenarios in which it currently lags behind is high, approaching that of the gold-standard manual evaluation metric, the Pyramid Method.

\section{Related Work}
\label{sec:related_work}
By far the most popular automatic methods for evaluating the content of a summary do so by comparing the tokens of the candidate and the reference.
The de facto metric ROUGE \citep{Lin2004} calculates a precision and recall score on the summaries' lexical overlap.
Recent methods BERTScore \citep{Zhang2019} and MoverScore \citep{Zhao2019} instead compare tokens based on the similarity of their contextual word embeddings.

Because these text overlap metrics do nothing to specifically measure how much information is common between two summaries, their scores are polluted by spurious matches between tokens that do not express the same information.
In contrast, QA-based evaluation metrics \emph{do} directly compare summaries based on their information.

The gold-standard for manually comparing two summaries' information overlap is the Pyramid Method \citep{Nenkova2004}.
It uses a domain-expert to identify spans of text between the candidate and reference summaries that express the same information, known as summary content units (SCUs).
Because the Pyramid Method's final score is calculated exclusively on the number of common SCUs, it is a purely information-based evaluation.

While there have been efforts to crowd source the Pyramid Method \citep{Shapira2019}, fully automatic approximations PEAK \citep{Yang2016} and PyrEval \citep{Gao2019} have also been proposed, with PyrEval reporting the best performance.
PyrEval identifies and matches SCUs by decomposing sentences into clauses, then calculating the similarity of the clauses based on their phrase embeddings.
This style of metric has been met with less success than text overlap metrics.

Several recent works also use QA to evaluate summaries.
\citet{narayan-etal-2018-ranking} use QA as part of a human evaluation to measure how much important document information was maintained by the summary.
FEQA \citep{Durmus2020} and QAGS \citep{Wang2020} automate evaluating the faithfulness of a summary.
Faithfulness and content quality are related, yet distinct, concepts.
Quality is a measure of whether the summary contains the correct information, whereas faithfulness measures whether the information is consistent with the input, regardless of its importance.
FEQA and QAGS compare summaries to the input documents, whereas we compare summaries to references.
Because the datasets used in our experiments are extractive summaries or have relatively high faithfulness ratings \citep{Fabbri2020}, we assume faithfulness is not an issue for simplicity.

Then, the most closely related work to ours is \citet{Eyal2019}, who also use QA to evaluate the content of summaries via their metric APES.
They create fill-in-the-blank questions by removing named entities from the reference summary and use a reading comprehension model to predict which entity was removed using the candidate summary.

There are several differences between their work and ours.
Our proposed metric QA\-Eval is more general than APES because QA\-Eval asks and answers questions about noun phrases, whereas APES is restricted to named entities.
APES may fail to accurately score summaries which do not have a sufficient number of named entities. 
Then, our evaluation of QA\-Eval is more comprehensive:
The experiments in \citet{Eyal2019} were limited to evaluating APES on 8 input instances from TAC'11,\footnote{\url{https://tac.nist.gov/}} whereas our experiments are run on 92 instances from benchmark content quality datasets TAC'08 and '09 as well as 100 instances from the CNN/DailyMail dataset \citep{Nallapati2016,Fabbri2020}.
Since our evaluation is more comprehensive and we demonstrate our metric has a high upper-bound performance, we believe it is a more convincing argument that QA-based metrics are a promising direction of future research.
Further, we perform an extensive evaluation on the individual components of the metric.
We compare our metric's performance to APES' in \S\ref{sec:overall} and \S\ref{sec:apes}.

\section{QA-Based Evaluation}
The standard line of research for evaluating the content quality of a summary is based on comparing the text of a candidate summary to a reference summary.
Metrics that follow this approach include ROUGE, Basic Elements \citep{Hovy2006}, Auto\-Summ\-ENG \citep{Giannakopoulos2008}, METEOR \citep{Denkowski2014}, BERT\-Score \citep{Zhang2019}, Mover\-Score \citep{Zhao2019}, and many more.

It is desirable to evaluate a summary based on the quality of the summary's information.
For reference-based metrics, this means measuring the overlap in information between the candidate and reference summary.
However, there is evidence that suggests text overlap metrics do not successfully accomplish this \citep{Deutsch2020c}.
They match tokens which do not express the same information and end up comparing the similarity of two summaries based on the topics they discuss.

We argue that a much better method of comparing the information content of two summaries is through QA.
In an ideal QA-based evaluation framework, all of the reference summary's information is represented by a set of QA pairs, and the candidate summary's recall of this information is measured by answering the questions against the candidate.
The questions should only be answerable if the information necessary to answer them is present in the candidate.
Therefore, this approach is fundamentally different from text overlap methods because it explicitly measures how much of the reference's information is contained in the candidate.

While we cannot yet achieve this ideal QA-based metric (our QA-based representations may be incomplete, our QA models are imperfect, etc.), we next propose a specific instantiation of this framework that represents our best effort at reaching this goal with today's state-of-the-art models.

\subsection{QAEval}
\label{sec:pipeline}
At the core of this work is a reference-based summarization evaluation metric that estimates the content quality of a summary, which we call QAEval.
The metric represents the information of a reference summary by a set of question-answer pairs that are automatically generated from the reference.
Then, QA\-Eval estimates how much of this information is in a candidate summary by using a learned QA model to answer the questions against the candidate.
The predictions from the QA model are verified as correct or incorrect, then the final score of the metric calculates what proportion of the questions were answered correctly.

Below, we describe the individual steps of the evaluation metric in more detail.
Then, each component of QAEval is analyzed individually in Sections~\ref{sec:answer_selection}, \ref{sec:question_generation}, and \ref{sec:question_answering} in order to identify any performance bottlenecks, followed by an overall evaluation of the metric in Section~\ref{sec:overall} and a reproduction of the experiments of \citet{Eyal2019} in Section~\ref{sec:apes}.

\paragraph{Answer Selection}
The first step in generating questions from the reference summary is to pick a set of phrases that represents answers to questions that will later be generated.
The answers should be chosen such that they will generate questions that cover as much of the information of the summary as possible.
We evaluate how much semantic content is represented by several different answer selection strategies in \S\ref{sec:answer_selection}.

\paragraph{Question Generation}
Once the answers have been selected, a learned model is used to generate a question for each answer.
The input to the question-generation model is a sentence which contains an answer phrase that is demarcated by special tokens.
The output is a question which is answerable by that phrase.

Following \citet{Durmus2020}, the generation model is a fine-tuned BART model \citep{Lewis2020} trained on 55k human-written question-answer pairs collected by \citet{Demszky2018}.
The quality of the generated questions and the impact of using model-generated questions instead of human-written questions on downstream correlations is measured in \S\ref{sec:question_generation}.

\paragraph{Question Answering}
Given a set of QA pairs generated from the reference summary, a QA model is used to answer the questions against the candidate summary.
Since there are no summarization datasets with labeled QA pairs, the QA model must be trained on a different dataset.
Further, because it is almost always the case that the candidate summary will not contain some reference summary information, it is necessary for the model to decide whether a question is answerable to reduce noise from spurious answers.
Therefore, the QA model is trained on the SQuAD 2.0 dataset \citep{Rajpurkar2018} which contains unanswerable questions.

The QA model is a pre-trained ELECTRA-Large model \citep{Clark2020} fine-tuned on SQuAD 2.0.
The input to the model is the candidate summary and a question.
The output is a span of text which contains the answer or a null string if the question is not answerable, depending on which is more probable under the model.
We estimate the answering performance of the QA model on the summarization data and estimate the improvement in downstream correlations that would be expected if the QA model had human-level performance in \S\ref{sec:question_answering}.

\paragraph{Answer Verification \& Scoring}
Finally, once the QA model has output predictions for all of the questions generated from a reference summary, they are verified as being correct or incorrect with respect to the ground-truth answers that were used to generate the questions.
We employ the two standard answer verification methods used by SQuAD, exact match (EM) and F$_1$ \citep{Rajpurkar2016}.
If the QA model outputs the null string, the score for that answer is 0.
We estimate whether these imperfect answer comparison strategies negatively impact downstream correlations in \S\ref{sec:question_answering}.

Finally, the metric produces two final scores that are the total EM and F$_1$ scores divided by the number of questions, thus calculating the proportion of questions answered correctly.
If multiple reference summaries are available, the scores are macro-averaged.
We refer to the metrics as QAEval-EM and QAEval-F$_1$.

\section{Experimental Methodology}
\label{sec:methodology}
We briefly review the experimental methodology that is used to evaluate metrics.

Evaluation metrics are used to estimate some property of a summary that is difficult to directly measure, such as the quality of its content.
In order to estimate how well the metric approximates the desired property (i.e., evaluate the evaluation metric), a set of summaries that have been annotated by human judges for that property is scored by the metric, and then the correlation between the two sets of scores is calculated.
The summaries are typically the outputs from multiple summarization models for the same set of inputs.

There are two standard ways to calculate correlations in the summarization literature: summary-level and system-level.
Assume $x^j_i$ and $y^j_i$ are two scores of metrics $X$ and $Y$ for the summary output by system $i \in \{1, \dots, N\}$ on input $j \in \{1, \dots, M\}$.
The summary-level correlation is calculated between the scores for each \emph{summary} for the same input, then averaged across inputs:
{
\begin{equation*}
    \rho_\textsc{Sum} = \frac{1}{M} \sum_j \textsc{Corr}\left(\left\{\left(x^j_i, y^j_i\right)\right\}_{i=1}^N\right)
\end{equation*}
}where $\textsc{Corr}(\cdot)$ calculates some correlation coefficient, typically Pearson $r$, Spearman $\rho$, or Kendall $\tau$.
Summary-level correlation measures how similarly $X$ and $Y$ rank summaries per-input. 
In contrast, the system-level correlation is calculated between the scores for each \emph{system} (typically the average score across the inputs):
\begin{gather*}
    \rho_\textsc{Sys} = \textsc{Corr}\left(\left\{\left(\frac{1}{M}\sum_j x^j_i, \frac{1}{M}\sum_j y^j_i\right)\right\}_{i=1}^N\right)
\end{gather*}
It measures how similarly $X$ and $Y$ rank summarization systems.

In this work, we examine how well evaluation metrics estimate the content quality of a summary using three English summarization datasets: the benchmark TAC'08 and '09 datasets \citep{Dang2008,Dang2009} as well as the subset of the CNN/DM dataset \citep{Nallapati2016} which was annotated by \citet{Fabbri2020}.

The TAC datasets consist of 48/44 multi-document summarization instances, each with 4 reference summaries written by human annotators.
Domain-expert judges rated the summaries output by 58/55 extractive models for each input on a scale of 1 to 5 based on how well they respond to an information need included in the task description.
Each summary is also assigned a Pyramid Score \citep{Nenkova2004} using a Pyramid constructed from the 4 reference summaries.
Our experiments on TAC calculate the correlations of the metrics to the responsiveness score for the 58/55 model summarizers and 48/44 instances.

The annotations provided by \citet{Fabbri2020} on the single-document summarization CNN/DM dataset score the outputs of 16 models across 100 instances.
The models are a mixture of extractive and abstractive approaches, and each instance has 1 reference summary.
\citet{Fabbri2020} collected relevance scores from 3 expert annotators that captures if the summary contains important content from the input document.
Our experiments report the correlation between the metrics' scores and the expert relevance judgments. 
\section{Answer Selection}
\label{sec:answer_selection}
In order for a QA-based evaluation metric to be successful, the QA pairs it uses to probe the candidate summary must represent a significant proportion of the reference summary's information.
Therefore, in this Section, we aim to understand how much information the QA pairs in QAEval do represent and whether that may limit the metric's performance.

\begin{figure}
    \centering
    \includegraphics[width=\columnwidth]{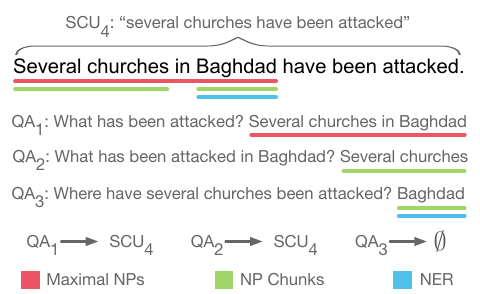}
    \caption{Example answers selected by the three strategies.
    The \emph{only} SCU marked by annotators for this sentence is SCU$_4$, which does not include information about the location of the attacks.
    Therefore, an answer selection strategy that chooses ``Baghdad'' enables generating a QA pair such as QA$_3$, which probes for information not included in the Pyramid annotation.
    }
    \label{fig:example_candidates}
\end{figure}

We explore three different answer selection strategies which pick phrases that are (1) named-entities, (2) noun phrase chunks, (3) or maximally-sized noun phrases.
The maximally-sized noun phrases in a sentence are identified by traversing the dependency tree down from the root until a noun is reached, then selecting the entire subtree for that noun.
Example answers selected by each strategy are presented in Figure~\ref{fig:example_candidates}.

Since there is no well-established method of measuring how much semantic content is represented by a set of QA pairs, we instead compare the content covered by the QA pairs to that of another semantic representation, the Pyramid Method SCUs (see \S\ref{sec:related_work} for details).
This approach allows us to compare answer selection strategies to a common point of reference as well as understand what types of information are represented by each formalism.

In order to compare the content covered by QA pairs and SCUs, each QA pair is manually mapped to an SCU based on whether the information that is being probed by the QA pair is included in the SCU description.
For instance, in Figure~\ref{fig:example_candidates}, QA$_1$ and QA$_2$ map to SCU$_4$ because they target what was attacked, which is included in the SCU description, whereas QA$_3$ would not because the SCU does not describe the location of the attacks.
This mapping allows us to calculate the proportion of QA pairs that map to some SCU, called the \emph{QA precision}, and the proportion of SCUs that are mapped to by some QA pair, called the \emph{SCU coverage}.

To ensure the generated questions are of high-quality, one of the authors manually wrote questions for every answer selected by each strategy for 20 reference summaries across 10 input document sets from TAC'08, totaling 801 questions.
The same author further mapped every QA pair to SCUs.
The results (averaged over reference summary) are presented in Table~\ref{tab:candidate_strategy}.

\begin{table}
    \centering
    \begin{adjustbox}{width=\columnwidth}
    \begin{tabular}{cccc}
        \toprule
        \bf Strategy & \bf Avg \#QAs & \bf QA Precision & \bf SCU Coverage \\
        \midrule
        NER & 11.7 & 83\% & 57\% \\
        NP Chunks & 28.8 & 79\% & 91\% \\
        Max. NPs & 17.3 & 82\% & 77\% \\
        \bottomrule
    \end{tabular}
    \end{adjustbox}
    \caption{The NP chunks answer selection strategy covers 91\% of the information represented by the Pyramid Method (SCU Coverage) with 21\% of the questions representing new information.
    From this, we conclude that the QA pairs generated from selecting noun chunk answers provides a semantic representation of the reference summary with very high-coverage.
    }
    \label{tab:candidate_strategy}
\end{table}

The most significant result we find is that the NP chunks strategy covers 91\% of the semantic information included in the Pyramid Method, with an additional 21\% of the questions targeting new information the Pyramid Method does not represent.
The other two strategies have much lower SCU coverages, likely because they result in fewer generated questions since their QA precisions are approximately equal to that of NP chunks.

This result is very promising for QA-based evaluation metrics because it indicates that the QA pairs cover nearly all of the information that is used by the Pyramid Method, the best-performing manual content quality evaluation.
Further, they even cover information the Pyramid Method does not, suggesting the potential for even better downstream correlations.
Therefore, we conclude that the information represented by the QA pairs generated from selecting noun chunk answers is unlikely to be a factor which limits QAEval's performance, and we subsequently use that selection strategy for the rest of our experiments.

\paragraph{Comparing QA Pairs \& SCUs}
Upon comparing the information that is represented by one formalism and not the other, there are some key differences.
The QA pairs miss information represented by nominal and adjectival modifiers because that information is contained within the answer noun phrase.
For instance, for sentence
\emph{[A Turkish novelist] was arrested}, the question asks about who was arrested, and not about the nationality of the novelist, which the SCUs do include.

In contrast, the SCUs often miss specific details and generalize over information that the QA pairs do not.
For instance, in Figure~\ref{fig:example_candidates}, although the SCUs do represent that the church attacks happened, it does not include information about their location, whereas this information is targeted by the QAs pairs.
\section{Question Generation}
\label{sec:question_generation}
An ideal question generation model should generate questions that are high enough quality that they do not impact the overall performance of the metric.
In this Section, we compare questions generated by the learned model to expert-written questions, both empirically and extrinsically through downstream correlations to human judgments.

\begin{figure}
    \centering
    \includegraphics[width=\columnwidth]{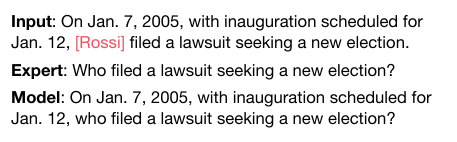}
    \caption{A typical example of expert-written and model-generated questions answerable by the phrase in red.
    The model questions are often significantly more verbose than the expert questions, typically copying the majority of the input sentence.
    }
    \label{fig:example_questions}
\end{figure}

\paragraph{Empirical Analysis}
Upon comparing the expert-written questions from \S\ref{sec:answer_selection} to model-generated questions for the same set of answers, we observe that 
a major difference between questions written by an expert versus a model is the level of verbosity.
The model-generated questions often copy most of the input sentence over to the question, including parts of the sentence which may not be relevant to answering the question.
In contrast, the questions written by an expert are more concise and remove the irrelevant details.
Examples of this difference can be seen in Figure~\ref{fig:example_questions}.

Despite the verbosity, nearly all of the model-generated questions are understandable to the authors.
However, because they are rather formulaic, the questions sometimes sound unnatural and could be confusing to a layman.
We did not find any examples in which the answer was included in the question.

\paragraph{Downstream Correlation}
Ideally, a QA-based evaluation metric would use an expert to write the questions to ensure they are all high-quality.
Unfortunately, this does not scale and is very expensive and time consuming, so the questions must be model-generated.
However, it is important to quantify any drop in performance caused by generating questions from a model rather than a domain-expert to understand the impact of using a less-than-ideal approach.

In order to measure any potential drop in performance, we compared the downstream correlations of the QA-based metrics to responsiveness judgments when using expert-written and model-generated questions.
In both cases, the question-answering component was done using the learned model described in \S\ref{sec:pipeline}.

\begin{figure}
    \centering
    \includegraphics[width=\columnwidth]{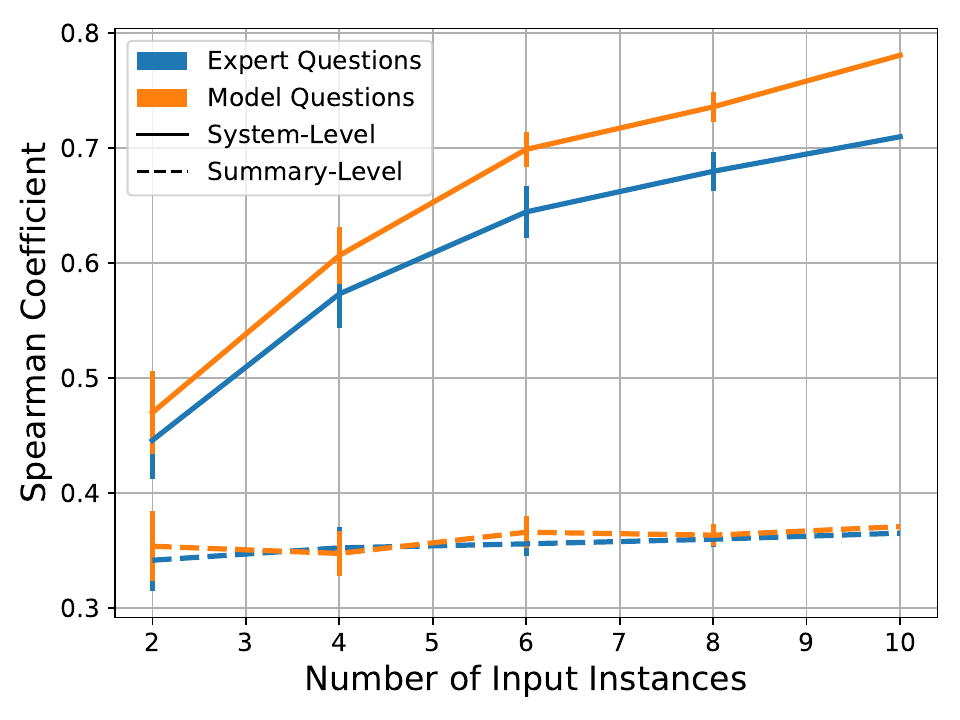}
    \caption{
    A comparison of the correlations of QAEval-F$_1$ on a subset of TAC'08 using expert-written and model-generated questions.
    Each point represents the average correlation calculated using 30 samples of $\{2,4,6,8,10\}$ instances, plotted with 95\% error bars.
    System-level correlations were calculated against the summarizers' average responsiveness scores across the entire TAC'08 dataset.
    We hypothesize the model questions perform better due to their verbosity, which causes more keywords to be included in the question that the QA model can match against the summary.}
    \label{fig:downstream_question_correlation}
\end{figure}

This experiment was performed on the subset of the TAC'08 dataset for which we collected expert-written questions (see \S\ref{sec:answer_selection}).
That is, the summaries from 58 different systems across 10 input instances with 2 references each were scored using the two setups, and the respective correlations were computed.\footnote{Since we do not have expert-written questions for all 4 references across all 48 input clusters, these results are not strictly comparable to later experiments (e.g., \S\ref{sec:overall}).}
We further simulated having a smaller number of input instances by downsampling the data to observe any emerging trends.
The results are plotted in Figure~\ref{fig:downstream_question_correlation}.

The downstream summary-level correlations appear near-identical between the two approaches.
However, surprisingly, the model-generated questions appear to result in better downstream correlations at the system-level than the expert-written questions.
As soon as around 6 input instances are available, the two curves separate from each other's margins of error, with the model-generated questions clearly trending with a Spearman correlation of at least 0.05 higher.

It is not clear from examining the data why this is the case;
There is no clear pattern that emerges which could explain why the model-generated questions result in higher correlations.
Our best hypothesis is that the verbosity of the generated questions helps the QA model by including more keywords that can be matched against the summaries to find an answer.

From these unexpected results, we can conclude that the model-generated questions do not harm the downstream correlations of QAEval at either the summary- or system-levels.
The rest of the experimentation in this paper will only use model-generated questions.
\section{Question Answering \& Verification}
\label{sec:question_answering}
The task of the QA model and answer verification step are to determine whether a question is answerable against a summary, predict an answer if it is, then compare the prediction to the ground-truth answer to determine if it is correct.
In this Section, we evaluate the performance of both components on the summarization data, first by calculating the QA performance (\S\ref{sec:qa_squad}) and then by estimating the downstream correlation of QAEval if both components had human-level performance (\S\ref{sec:expert_answers}).

\subsection{Question-Answering Model Performance}
\label{sec:qa_squad}
Since the QA model is trained on Wikipedia articles in the SQuAD 2.0 dataset and used to answer questions generated from the summarization data, it is expected that the QA performance on the summarization data will be worse than on the original training data due to the domain shift.

In order to quantify the size of such a drop, one of the authors manually answered 2.9k generated questions from 20 reference summaries across 10 input clusters against 4 different summarizers on TAC'08 and 2.3k generated questions across 10 input documents against all 16 summarizers on CNN/DM.
For each question and summary pair, it was first determined whether the summary contained the answer to the question, then if it did, a span of text was selected as the answer.
Then, the selected answer was later manually verified as correctly or incorrectly answering the question.

We compare the QA model's ability to both identify if a question is answerable and to select the correct answer if one exists separately on SQuAD 2.0 and the summarization datasets.
This is done to measure any performance decrease on each problem in isolation.
We calculated the F$_1$ score on the model's predictions on whether the question is answerable, plus the standard SQuAD EM and F$_1$ metrics on only the subset of QA pairs for which the ground-truth and model agree that the question is answerable.
We do not want to measure the quality of the predicted answer if the question is not answerable or the model outputs no answer.

\begin{figure}
    \centering
    \includegraphics[width=\columnwidth]{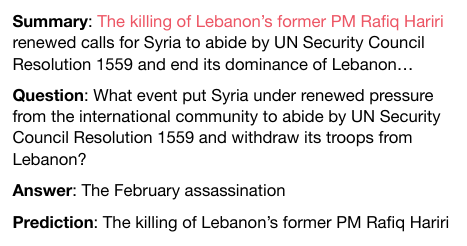}
    \caption{An example correct answer predicted by the model that is scored poorly by the EM or F$_1$ QA metrics (both would assign a score of 0 or near 0).
    This occurs because the answer and prediction are drawn from two different summaries, and the same event is referred to in different ways in each one.
    }
    \label{fig:answer_verification}
\end{figure}

In addition to EM and F$_1$, we also report the correct answer accuracy according to the human annotator.
EM and F$_1$ are imperfect answer comparison strategies because they may fail to identify an answer as correct if it is a paraphrase of the ground-truth.
Unlike SQuAD, the ground-truth answer and model prediction come from different source texts, increasing the likelihood that both answers will be expressed differently (see Fig.~\ref{fig:answer_verification}).
Comparing the human annotator accuracy to EM and F$_1$ will quantify how well the automatic answer verification methods work on the summarization data.

\begin{table}
    \centering
    \begin{adjustbox}{width=\columnwidth}
    \begin{tabular}{cccccc}
        \toprule
        \multirow{2}{*}{\bf Dataset} & \multirow{2}{*}{\bf \%IsAns} & \multirow{2}{*}{\bf IsAns-F$_\mathbf{1}$} & \multicolumn{3}{c}{\bf Given IsAns} \\
        \cmidrule{4-6}
        & & & \bf EM & \bf F$_1$ & \bf Acc \\
        \midrule
        SQuAD 2.0 & 50.0\% & 92.0 & 88.0 & 94.5 & - \\
        TAC'08 & 14.2\% & 52.4 & 30.5 & 47.2 & 84.3 \\
        CNN/DM & 36.3\% & 75.3 & 33.2 & 52.0 & 86.3 \\
        \bottomrule
    \end{tabular}
    \end{adjustbox}
    \caption{The QA performance on the summarization datasets drops significantly compared to its performance on SQuAD, especially for TAC'08.
    This is expected due to the domain shift, however we suspect the drop is smaller for CNN/DailyMail because the generated and reference summaries are far more similar than for TAC, thus making it easier to answer questions.}
    \label{tab:squad_results}
\end{table}

The results are presented in Table~\ref{tab:squad_results}.
In general, the QA performance drops for both datasets, but the decrease is more extreme for TAC'08.
Specifically, we see that the drops in IsAns-F$_1$ are significant, amounting to decreasing by nearly 40 points from 92.0 on SQuAD to 52.4 on TAC'08 and almost 17 points to 75.3 on CNN/DM.
This result indicates that identifying if a question is answerable is very challenging for the model, especially on TAC.

The EM and F$_1$ results across datasets also see a rather significant drop of around 47-58 points for TAC and 42-55 for CNN/DM, pointing to a much worse answering performance by the model when the model correctly predicts that an answer exists.
However, the accuracy according to the human annotations is closer to the performance on SQuAD, implying the actual drop in performance is actually not as significant.
The discrepancy between the EM/F$_1$ scores and human accuracy judgments means the model's predictions are frequently correct, but EM and F$_1$ fail to identify them as such in a significant number of cases, thereby implying they are noisy answer verification methods.
This problem has been observed for QA models before \citep{Wang2020,Chen2020}, but the issue seems particularly apparent for when the answer and QA model's prediction come from different texts.

We suspect that the QA model fares better on CNN/DM than TAC because the CNN/DM generated summaries are far more similar to the reference summaries than those in TAC.
This is likely due to several factors:
(1) The CNN/DM task is in some sense easier than the TAC task.
The lead-3 baseline is very strong, so the models can more easily generate high quality summaries;
(2) The models included in the annotation are more recent state-of-the-art models compared to those from TAC and are likely better summarizers;
(3) The task is single-document, so the information in the reference and generated summaries is more likely to be expressed the same way.

Since the two summaries being compared are similar to each other, the generated questions have a large token overlap with the target summary.
This likely results in the QA model being more effective at identifying when an answer exists in the summary and then subsequently correctly identifying it.
We expect this result to hold for other popular single-document summarization datasets.

From this experiment, we conclude that identifying whether a question is answerable and subsequently verifying whether the QA model's prediction is correct are potential performance bottlenecks QA\-Eval.

\subsection{Human-Level Performance Comparison}
\label{sec:expert_answers}
After identifying QA and answer verification as potentially problematic for QAEval's performance, we now estimate the size of any potential drop in downstream correlation compared to using human-level performance for both of those components.

Using the same human-annotated QA pairs from the previous Section, we calculated the summary-level correlations of QAEval when it uses either human annotations for the QA model, human annotations for the answer verification, or both.
The correlations for these QAEval variants and several other metrics (discussed in \S\ref{sec:related_work}) are in Table~\ref{tab:expert_answering}.

Since this experiment only uses a relatively small amount of data, none of the correlations differ by a statistically significant margin, so coming to definitive conclusions is difficult.
However, some trends do emerge from the data.

For TAC'08, QAEval is competitive to the other evaluation metrics when it uses a learned QA model and F$_1$ verification.
Then, human-level performance for both QA and answer verification provide large improvements in the downstream correlations, both independently and when combined.
For instance, human QA annotations improve QAEval on TAC by 0.12 and 0.14 Pearson with F$_1$ and human verification, respectively.
Human annotations for answer verification improve QAEval with model and human QA components by 0.17 and 0.29 Spearman, respectively.
When both components use human annotations, the correlations are significantly better than any of the other automatic metrics and approach those of the Pyramid Method.

The results on CNN/DM are less clear.
There is no obvious pattern in the data and all of the model/human combinations result in roughly the same performances.
We suspect that because the drop in QA performance is less significant (\S\ref{sec:qa_squad}), the differences in model and human-level QA performance is not reflected on CNN/DM as it is on TAC.
Further, we empirically observed that the content of this dataset's summaries are more similar in content across models than the TAC summaries, making them harder to rank (as demonstrated by the lower correlations), which would also introduce more variance to the correlations.

Overall, this is a promising result for the future potential of QA-based evaluations, especially for more complex multi-document summarization tasks which are in some sense harder for metrics to evaluate than single-document summaries.
While the current summary-level results on both datasets may be competitive to other metrics, the metric's upper-bound performance is very high on TAC and is approaching the gold-standard manual evaluation, the Pyramid Method.

\begin{table}
    \centering
    \begin{adjustbox}{width=\columnwidth}
    \begin{tabular}{cccccccccc}
        \toprule
        \multicolumn{3}{c}{\multirow{2}{*}{\bf System}} & \multicolumn{3}{c}{\bf TAC'08} & & \multicolumn{3}{c}{\bf CNN/DM} \\
        \cmidrule{4-6} \cmidrule{8-10}
        \multicolumn{3}{c}{} & $r$ & $\rho$ & $\tau$ & & $r$ & $\rho$ & $\tau$ \\
        \midrule
        \multicolumn{3}{c}{Pyramid Score} & .63 & .69 & .65 & & - & - & - \\
        \multicolumn{3}{c}{ROUGE-1} & .27 & .27 & .26 & & .25 & .21 & .18 \\
        \multicolumn{3}{c}{ROUGE-2} & .34 & .40 & .38 & & .13 & .09 & .06 \\
        \multicolumn{3}{c}{ROUGE-L} & .20 & .22 & .21 & & .13 & .12 & .08 \\
        \multicolumn{3}{c}{ROUGE-SU4} & .29 & .22 & .22 & & .16 & .16 & .12 \\
        \multicolumn{3}{c}{MoverScore} & .42 & .28 & .28 & & .27 & .23 & .18 \\
        \multicolumn{3}{c}{APES} & .35 & .38 & .37 & & .08 & .09 & .07 \\
        \midrule
        \multicolumn{3}{c}{\footnotesize  \bf QAEval} & & & \\
        QA & & Verif. & & & \\
        \cmidrule{1-1} \cmidrule{3-3}
        Model & & F$_1$ & .31 & .28 & .26 & & .21 & .23 & .18 \\
        Human & & F$_1$ & .43 & .33 & .30 & & .15 & .14 & .12 \\
        Model & & Human & .44 & .45 & .42 & & .25 & .24 & .20 \\
        Human & & Human & .58 & .62 & .59 & & .22 & .21 & .17 \\
        \bottomrule
    \end{tabular}
    \end{adjustbox}
    \caption{Summary-level correlations calculated using 4 systems across 10 inputs on TAC and 16 systems across 10 inputs on CNN/DailyMail compared using answers from a model or a human and verifying if the answer is correct using F$_1$ or a human.
    Because the results are on a small sample of the dataset, the results are not statistically significant.
    However, the trend on TAC is that human-level performance greatly improves the results, approaching correlations equal to the Pyramid Method's.
    On CNN/DailyMail, we suspect the same trend does not appear because the QA model performs much better than on TAC.
    }
    \label{tab:expert_answering}
\end{table}
\begin{table*}[t]
    \begin{subtable}[h]{\columnwidth}
    \centering
    \begin{adjustbox}{width=\columnwidth}
    \begin{tabular}{cccccccc}
        \multicolumn{8}{c}{\bf TAC 2008} \\
        \toprule
        \multirow{2}{*}{\textbf{Metric}} & \multicolumn{3}{c}{\textbf{System-Level}} & & \multicolumn{3}{c}{\textbf{Summary-Level}} \\
        \cmidrule{2-4} \cmidrule{6-8}
        & $r$ & $\rho$ & $\tau$ & & $r$ & $\rho$ & $\tau$ \\
        \midrule
        Pyramid Score & .90 & .88 & .70 & & .59 & .59 & .50 \\
        \midrule
        ROUGE-1 & .79 & .80 & .60 & & \underline{.49} & \underline{.48} & \underline{.39} \\
        ROUGE-2 & .83 & .87 & .67 & & \underline{.48} & \underline{.48} & \underline{.39} \\
        ROUGE-L & .74 & .77 & .57 & & .46 & .45 & .36 \\
        ROUGE-SU4 & .80 & .83 & .63 & & \underline{.49} & \underline{.48} & \underline{.39} \\
        PyrEval & .81 & .79 & .59 & & .31 & .31 & .25 \\
        MoverScore & .83 & .82 & .63 & & \bf .50 & \bf .49 & \bf .40 \\
        APES & .74 & .82 & .60 & & .25 & .25 & .21 \\
        \midrule
        QAEval-EM & \bf .93 & \bf .91 & \bf .76 & & .33 & .33 & .27 \\
        QAEval-F$_1$ & .90 & \underline{.88} & .71 & & .46 & .45 & .36 \\
        \bottomrule
    \end{tabular}
    \end{adjustbox}
    \end{subtable}
    \hfill
    \begin{subtable}[h]{\columnwidth}
    \centering
    \begin{adjustbox}{width=\columnwidth}
    \begin{tabular}{cccccccc}
        \multicolumn{8}{c}{\bf TAC 2009} \\
        \toprule
        \multirow{2}{*}{\textbf{Metric}} & \multicolumn{3}{c}{\textbf{System-Level}} & & \multicolumn{3}{c}{\textbf{Summary-Level}} \\
        \cmidrule{2-4} \cmidrule{6-8}
        & $r$ & $\rho$ & $\tau$ & & $r$ & $\rho$ & $\tau$ \\
        \midrule
        Pyramid Score & .90 & .87 & .70 & & .59 & .57 & .48 \\
        \midrule
        ROUGE-1 & .83 & .78 & .60 & & \bf .54 & .47 & .38 \\
        ROUGE-2 & .76 & .84 & \underline{.67} & & .50 & \underline{.50} & \underline{.40} \\
        ROUGE-L & .82 & .72 & .54 & & \bf .54 & .47 & .37 \\
        ROUGE-SU4 & .77 & .81 & .63 & & .52 & .50 & .39 \\
        PyrEval & \underline{.86} & \underline{.82} & \underline{.64} & & .39 & .35 & .28 \\
        MoverScore & \underline{.82} & \underline{.80} & \underline{.63} & &  \underline{.51} & \bf .52 & \bf .42 \\
        APES & \bf .87 & \underline{.80} & \underline{.63} & & .41 & .35 & .28 \\
        \midrule
        QAEval-EM & .70 & \underline{.87} & \underline{.69} & & .42 & .38 & .30 \\
        QAEval-F$_1$ & .81 & \bf .89 & \bf .72 & & .50 & .45 & .36 \\
        \bottomrule
    \end{tabular}
    \end{adjustbox}
    \end{subtable}
    \caption{The Pearson $r$, Spearman $\rho$, and Kendall $\tau$ correlation coefficients calculated between the metrics' scores and expert responsiveness judgments on the TAC'08 (left) and TAC'09 (right) datasets.
    QA\-Eval has the highest system-level correlations, even better than the fully manual Pyramid Score, whereas the summary-level correlations are lower (EM) or competitive (F$_1$) with other metrics.
    We believe this supports our hypothesis that the QA model and answer verification are noisy (causing lower summary-level correlations) but average out to a high-quality metric given enough QA pairs (causing high system-level correlations).
    On TAC'09, the QA $r$ values are much lower because of an outlier, and $r$ is sensitive to outliers.
    If the outlier is removed, the $r$ values become 0.92 and 0.93 for EM and F$_1$.
    }
    \label{tab:responsiveness}
\end{table*}

\section{Overall Metric Analysis}
\label{sec:overall}
After analyzing each component of QAEval, we now turn to calculate the metric's correlations to human responsiveness/relevance judgments on TAC'08, '09, and CNN/DM (see \S\ref{sec:methodology} for more details about the experimental methodology; An additional experiment that varies the number of available references is included in Appendix~\ref{sec:num_ref_learn_curve}).
For this experiment, QA\-Eval uses the NP chunks answer selection strategy and learned question-generation and question-answering models and is therefore a fully automatic metric.

In addition to the QAEval correlations, we also report those of several baselines and state-of-the-art metrics, including the Pyramid Score, several variants of ROUGE, PyrEval, and MoverScore (which reports better correlations than BERTScore), and APES.
See \S\ref{sec:related_work} for descriptions of these metrics.
Results in bold are the highest among the automatic metrics.
Those underlined are statistically indistinguishable from the highest under a single-tailed permutation test for correlations with $\alpha = 0.05$ \citep{DeutschDrRo21}.

\paragraph{TAC'08 \& '09}
The correlations for TAC are presented in Table~\ref{tab:responsiveness}.
First, we see that the summary-level correlations for the QAEval metrics are lower than or comparable to some of the other automatic metrics.
For example, the TAC'08 Pearson's $r$ for QA\-Eval-EM is 0.33, whereas the $r$ values for QAEval-F$_1$ and ROUGE-2 are 0.46 and 0.48.
Given that the QA model and answer verification components introduce noise into the metric, this result is consistent with the analysis in \S\ref{sec:expert_answers} and unsurprising.

However, the system-level results are quite surprising.
The QAEval metrics achieve state-of-the-art system-level performance on nearly every correlation coefficient across both datasets, reaching correlations comparable to the Pyramid Method itself.
For instance, on TAC'08, QA\-Eval-EM has a Kendall's $\tau$ of 0.76 compared to 0.70 for the Pyramid Method and 0.67 for the next-highest automatic metric, ROUGE-2.
This pattern largely holds for TAC'09, with the exception of Pearson's $r$ due to an outlier.\footnote{
Once removed, the $r$ values are 0.92 and 0.93 for QAEval-EM and QAEval-F$_1$, higher than any other metric.
}

It is unexpected that QAEval should achieve both state-of-the-art system-level results and lower summary-level results simultaneously and that the system-level results are even better than the Pyramid Method's.

We believe the discrepancy between the summary- and system-level results can be explained by the number of questions that is used by each evaluation.
QAEval estimates the quality of an individual summary using around 110 questions.
In contrast, the system-level scores are based over 5,000 QA pairs across 48 or 44 instances.
We suspect that when QAEval's scores are averaged over such a large number of questions, the metric is able to overcome any noise introduced by the QA model or answer verification, resulting in a high-quality evaluation.
APES, the other QA-based metric, also exhibits a similar pattern, supporting this hypothesis.

Then, it is likely that QA\-Eval's system-level performance rivals the Pyramid Method's because the QA pairs probe for more semantic content than is represented by the SCUs (\S\ref{sec:answer_selection}).
The QA model and answer verification largely perform the same task as the Pyramid Method annotators: identify a span of text in the candidate summary which expresses a specific piece of information.
It is unlikely the models do this better than a human, even after the noise is averaged out across thousands of examples.
Therefore, it must be the case that the semantic representation of the QA pairs provides better coverage of the reference summary than the SCUs do, resulting in comparable overall performance.

\paragraph{CNN/DM}
The results on the CNN/DM dataset are shown in Table~\ref{tab:fabbri2020}.
Compared to TAC, the improvement in system-level correlations is significantly larger.
For instance, both QAEval variants achieve a system-level Spearman 0.91, whereas the next highest metrics APES and ROUGE-1 reach 0.73 and 0.62.
Unlike for TAC, the summary-level correlations are either higher or statistically indistinguishable from the other metrics.

\begin{table}
    \centering
    \begin{adjustbox}{width=\columnwidth}
    \begin{tabular}{cccccccc}
        \multicolumn{8}{c}{\bf \citet{Fabbri2020}} \\
        \toprule
        \multirow{2}{*}{\textbf{Metric}} & \multicolumn{3}{c}{\textbf{System-Level}} & & \multicolumn{3}{c}{\textbf{Summary-Level}} \\
        \cmidrule{2-4} \cmidrule{6-8}
        & $r$ & $\rho$ & $\tau$ & & $r$ & $\rho$ & $\tau$ \\
        \midrule
        ROUGE-1 & .61 & .62 & .50 & & \underline{.28} & \underline{.26} & \underline{.20} \\
        ROUGE-2 & .64 & .60 & .43 & & .23 & .19 & .14 \\
        ROUGE-L & .61 & .48 & .32 & & .21 & .18 & .14 \\
        ROUGE-SU4 & .62 & .56 & .38 & & .23 & .19 & .15 \\
        MoverScore & .56 & .54 & .42 & & \underline{.28} & .24 & .18 \\
        APES & .68 & .73 & \underline{.58} & & .10 & .09  & .07 \\
        \midrule
        QAEval-EM & .80 & \bf .91 & \bf .77 & & .23 & .23 & .19 \\
        QAEval-F$_1$ & \bf .82 & \bf \bf .91 & \bf .77 & & \bf .30 & \bf .29 & \bf .22 \\
        \bottomrule
    \end{tabular}
    \end{adjustbox}
    \caption{The QAEval metrics on the CNN/DailyMail annotations provided by \citet{Fabbri2020} achieve significantly higher correlations than the other authomatic metrics, likely due to the relatively good QA model performance on this dataset compared to on TAC.}
    \label{tab:fabbri2020}
\end{table}

We hypothesize that the improved performance on CNN/DM compared to TAC is due to the QA model's quality on this dataset.
In \S\ref{sec:qa_squad}, we demonstrated that the QA performance did drop on CNN/DM with respect the model's results on the SQuAD data, however that performance decrease was not nearly as large on CNN/DM as on TAC.
Since the QA model and answer verification are the performance bottlenecks and both suffer less on CNN/DM, the QAEval metrics achieve strong correlations.

This result is evidence to support that QAEval is a very effective metric for evaluating current state-of-the-art systems on today's popular summarization datasets.

\paragraph{Comparison to APES}
Across all three datasets, QA\-Eval achieves higher or comparable correlations than the other QA-based metric, APES, at both the summary- and system-levels.
We suspect this is due to at least two reasons.
First, their reading comprehension model likely has lower performance than the QA model used in QA\-Eval.
The QA\-Eval pretrained model leverages recent state-of-the-art models that use contextual word embeddings, which the model of \citet{Eyal2019} does not use.
Second, APES targets named entities in the summaries, which we demonstrated does not probe for as much information as using all noun phrases (\S\ref{sec:answer_selection}).
If the summaries do not contain a sufficient number of entities, APES may fail to accurately score it.

\paragraph{Overall}
Since the performance of QAEval using EM and F$_1$ is roughly equal at the system-level, but F$_1$ is clearly better at the summary-level, we recommend that future work which evaluates with QAEval use the F$_1$ variant.

Overall, since evaluation metrics are most commonly used in the summarization community to rank summarization systems, these experimental results suggest that QAEval is one of the most effective evaluation metrics to date.

\section{APES Experiments}
\label{sec:apes}
To further compare QAEval to APES, we reproduce some of the experiments reported by \citet{Eyal2019} and compare the results of the two metrics.

\subsection{TAC 2011 Comparison}
First, we compare the summary-level correlations of the two metrics and ROUGE to human judgments on a subset of the TAC'11 dataset.
TAC'11 contains extractive summaries produced by 51 models on 44 input document sets.
However, \citet{Eyal2019} evaluate on the 8 input document sets about ``Investigations and Trials'' for which there were a sufficient number of named entities.
This is because the QA model used by APES is only trained to predict named entities as answers.
Similarly to TAC'08 and '09, each summary has an overall responsiveness score and a Pyramid score that were annotated by domain experts.

\begin{table}
    \centering
    \begin{adjustbox}{width=\columnwidth}
    \begin{tabular}{cccccccc}
        \toprule
        & R1 & R2 & RL & RSU4 & APES & QA-EM & QA-F1 \\
        \midrule
        Pyr. & .73 & .73 & .70 & .74 & .61 & .47 & .61 \\
        Resp. & .62 & .65 & .60 & .63 & .50 & .46 & .56 \\
        \bottomrule
    \end{tabular}
    \end{adjustbox}
    \caption{Summary-level Pearson correlations of ROUGE, APES, and QAEval to overall responsiveness and the Pyramid Score on the 8 instances from TAC'11 that were used in \citet{Eyal2019}.
    These numbers differ from those reported by \citet{Eyal2019} because they directly calculate the correlation between the scores for all of the summaries across all instances (personal communication with the authors).
    This differs from the standard definition of the summary-level correlation, which calculates a correlation per input document set \citep{Louis2013}, then averages the correlations
    \citep[see $\rho_\textsc{sum}$ in \S\ref{sec:methodology}; ][]{Peyrard2017b,Zhao2019,BhandariGoAsLiNe20}.
    }
    \label{tab:tac2011}
\end{table}

Table~\ref{tab:tac2011} contains the summary-level Pearson correlations of ROUGE, APES, and QA\-Eval to the human judgments on the subset of TAC'11.
Although it is difficult to come to conclusions on this dataset due to its relatively small size, we observe that APES out-performs QA\-Eval-EM and under-performs QA\-Eval-F$_1$ using the responsiveness score as the ground-truth.
Using the Pyramid score as the ground-truth, APES and QA\-Eval-F$_1$ are equal.
However, both QA-based metrics are lower than the ROUGE variants, which is consistent with both APES and QA\-Eval achieving lower correlations than ROUGE on TAC'08 and '09 at the summary-level.
The APES correlations here are much higher on this subset of TAC'11 than on the whole of TAC'08 and '09, supporting that its performance is higher when the summaries have a sufficient number of named entities.

\subsection{Complementary Signals}
\label{sec:pairwise_correlations}
\begin{figure}
    \centering
    \includegraphics[width=\columnwidth]{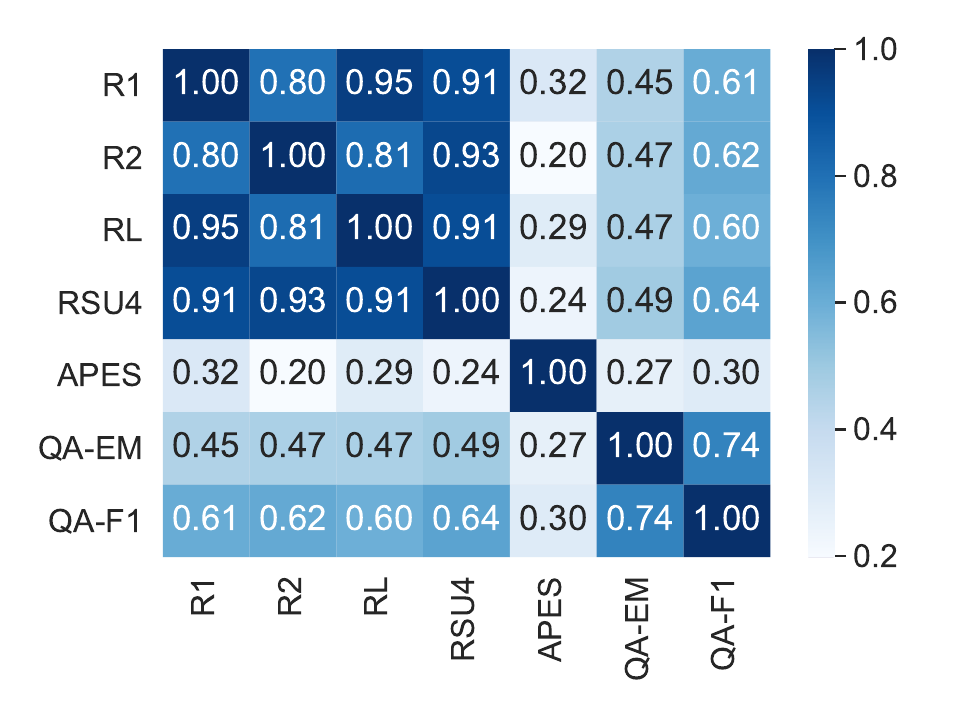}
    \caption{The Pearson correlations between the scores of several ROUGE variants, APES, and QAEval variants on TAC'08.
    The results support similar findings of \citet{Eyal2019}, namely that the ROUGE metrics are highly correlated to each other but have low correlation to the QA-based metrics, suggesting the two types of metrics offer complementary signals.
    }
    \label{fig:pairwise_correlations}
\end{figure}

Then, \citet{Eyal2019} demonstrate that APES and ROUGE are less correlated to each other than ROUGE variants are to themselves, suggesting they offer complementary signals of summary quality.
In Table~\ref{fig:pairwise_correlations} we show the Pearson correlations between several different variants of ROUGE, APES, and QAEval on the TAC'08 summaries.

Our results suggest similar conclusions to \citet{Eyal2019}.
Specifically, each of the ROUGE variants is very highly correlated to each other ($\geq .80$), whereas the correlations to the QA-based metrics are lower ($\approx .47$ for QAEval-EM, $.62$ for QAEval-F$_1$, and $.26$ for APES).
Interestingly, APES and QA\-Eval are as correlated to each other as APES is to ROUGE.
We hypothesize that because the QA models are trained on different corpora (CNN for APES versus Wikipedia for QA\-Eval), they learn different signals to answer questions and are more effective at scoring different summaries.
Future work could explore combining lexical overlap and QA-based methods into a single metric.

\section{Discussion}

\paragraph{Limitations}
Overall, QA\-Eval is limited by its dependence on using predicate-argument relations throughout each component of the metric.
QA\-Eval represents summaries with QA pairs that target nouns as answers, which is insufficient for representing all of the summary's information (as pointed out in \S\ref{sec:answer_selection}).
The question generation model is limited to producing questions that reason about the arguments of predicates and cannot generate more abstract questions (e.g., \emph{What types of conflict have there been?} for Fig.~\ref{fig:example_candidates}).
Likewise, QA models trained on SQuAD-style questions can only reason about matches between predicate-argument relations and cannot answer more abstract questions even if the generation model could produce them.

Because of this dependence on predicate-argument relations, any similarity between summaries that cannot be represented by matching predicates and arguments can also not be captured by QA\-Eval.
Although this does not appear to be an issue in our experiments, we anticipate that using generation and answering models which are capable of a more sophisticated level of reasoning will be necessary in the future.

\paragraph{QA-Based versus Text Overlap}
Although QA\-Eval has superior or comparable system-level correlations on the datasets included in our experimentation, it still lags behind text overlap-based method ROUGE at the summary-level in some settings.
Therefore, we do not recommend completely replacing text overlap metrics with QA\-Eval, nor do we believe that this should be done even if a QA-based metric achieves summary-level parity.

Both \citet{Eyal2019} and our work clearly show evidence that QA-based metrics provide a summary quality signal that is complementary to ROUGE (\S\ref{sec:pairwise_correlations}), yet both ROUGE and QA\-Eval achieve strong correlations in our experiments.
The quality signals captured by these metrics are clearly both valuable and different.
Evaluating a summarization model with only one type of metric would miss out on summary quality signals captured by the other.
Therefore, we recommend future work use both a text overlap metric as well as a QA-based metric to evaluate their summarization models.

\section{Conclusion}
In this work, we proposed a QA-based evaluation metric called QA\-Eval.
We demonstrated that QAEval already achieves state-of-the-art system-level correlations, and we estimate its upper-bound summary-level performance on multi-document summaries is quite high.
Through a careful analysis of each component of QA\-Eval, we identified that the performance bottlenecks are both the QA model and verifying whether or not the QA model's predicted answer is correct.
We believe that these results are strong evidence that QA-based evaluation metrics are a promising direction for future research on summarization evaluation.

\section*{Acknowledgments}
The authors would like to thank Annie Louis, Shashi Narayan, Yang Gao, Fei Liu, and the anonymous TACL reviewers for their very detailed and helpful feedback on our paper, which we feel significantly strengthened our work.

This work was supported by Contracts FA8750-19-2-1004 and FA8750-19-2-0201 with the US Defense Advanced Research Projects Agency (DARPA). Approved for Public Release, Distribution Unlimited. The views expressed are those of the authors and do not reflect the official policy or position of the Department of Defense or the U.S. Government.

This research is supported by a Focused Award from Google.

\bibliography{summarization,tacl,ccg,cited}
\bibliographystyle{acl_natbib}

\appendix

\clearpage
\section{Number of Available References}
\label{sec:num_ref_learn_curve}
Previous work has argued that multiple reference summaries are necessary for metrics to achieve stable correlations to ground-truth annotations, especially at the summary-level \citep{Nenkova2004,Louis2013}.
Since the TAC datasets provide four reference summaries per input, we are able to measure how much benefit the additional references provide by simulating having fewer references.

\begin{figure}[t]
    \centering
    \includegraphics[width=\columnwidth]{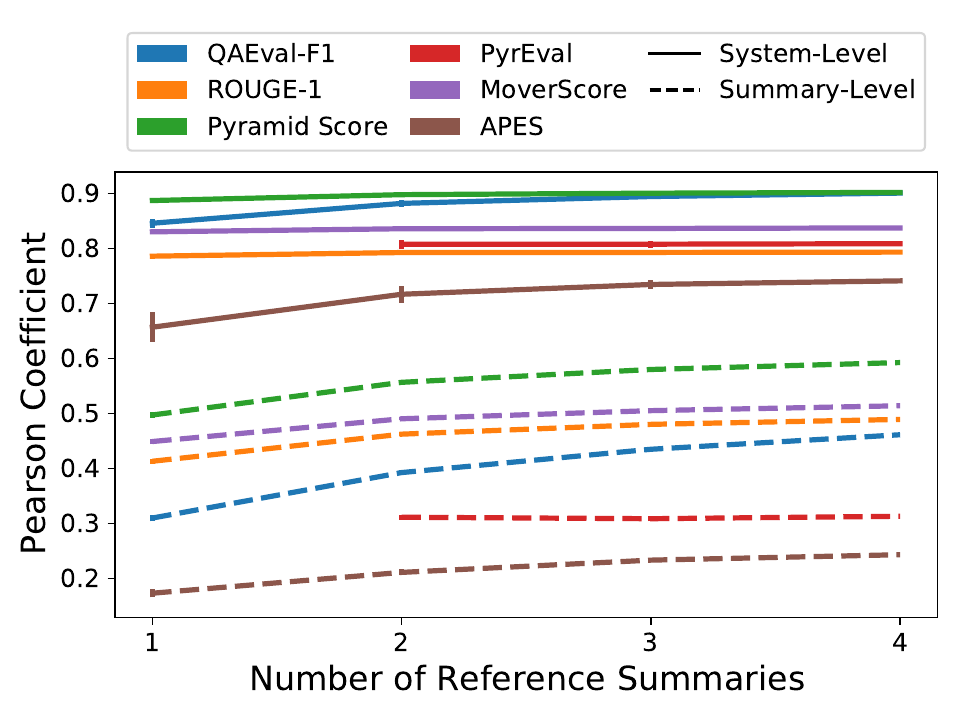}
    \caption{The system- and summary-level Pearson correlations as the number of available reference summaries increases.
    95\% confidence error bars shown, but may be too small to see.
    PyrEval is missing data because the official implementation requires at least two references.
    Even with one reference summary, QAEval-F$_1$ maintains a higher system-level correlation than ROUGE.}
    \label{fig:num_ref_curve}
\end{figure}

In order to simulate only having one reference summary, for each input document set from TAC'08, we randomly sample one reference, score all of the peer summaries against only that reference, and calculate the correlation to the responsiveness scores.
We collect 30 such samples and report the average correlation.
This procedure is also repeated for two and three references to understand the impact of each additional reference.
The results are plotted in Figure~\ref{fig:num_ref_curve}.

At the system level, the Pearson correlations are largely the same when the metrics are provided with one or four references.
This is in agreement with \citet{Louis2013}, who show system-level results are relatively stable with either one or four references.
Among the metrics, the QA-based metrics see the largest improvement in performance with adding additional references.
QAEval-F$_1$ increases from 0.85 with one reference compared to 0.90 with four.
Despite its drop in performance with one reference, QAEval-F$_1$ is still better than ROUGE even with four references at 0.79.
APES improves from 0.66 to 0.74.

When the metrics are compared at the summary level, it is clear that the correlations for each metric are less stable.
Nearly all of the metrics greatly benefit from additional references:
Pyramid Score improves by 0.09 (+19\%), ROUGE by 0.08 (+18\%), and QAEval-F$_1$ by 0.15 (+49\%).
The large improvement by QAEval-F$_1$ is further evidence that the noisy question-answering model averages out to a high-quality responsiveness estimator when provided with a large number of QA pairs.

Overall, QAEval does incur a significant performance drop at the summary-level, but since most comparisons of summarization systems are done at the system-level, it does not appear that having multiple references per input is necessary for good results.

\end{document}